\pdfoutput=1

\pdfoutput=1

\documentclass[11pt]{article}
\usepackage[colorinlistoftodos]{todonotes}

\usepackage[]{coling}

\usepackage{booktabs}
\usepackage{times}
\usepackage{latexsym}
\usepackage{graphicx}
\usepackage{url} 
\usepackage[T1]{fontenc}

\usepackage[utf8]{inputenc}

\usepackage{microtype}

\usepackage{inconsolata}

\usepackage{multirow}
\usepackage{rotating}
\usepackage{todonotes}
\usepackage{multicol}

\usepackage[usestackEOL]{stackengine}
\usepackage{tabulary}
\usepackage[maxfloats=30,morefloats=12]{morefloats}
\usepackage{booktabs}
\usepackage{float,lscape}
\usepackage{longtable}
\usepackage{pdflscape}
\usepackage{tabularx}
\usepackage{multirow}
\usepackage{bigstrut}
\usepackage{fdsymbol}

\title{What Makes Cryptic Crosswords Challenging for LLMs?}

\author{Abdelrahman Sadallah \qquad  Daria Kotova \qquad Ekaterina Kochmar \\
Department of Natural Language Processing, MBZUAI \\
\texttt{\{abdelrahman.sadallah, daria.kotova, ekaterina.kochmar\}@mbzuai.ac.ae 
	}
}

\begin{document}
\maketitle
\begin{abstract}
Cryptic crosswords are puzzles that rely on general knowledge and the solver's ability to manipulate language on different levels, dealing with various types of wordplay. Previous research suggests that solving such puzzles is challenging even for modern NLP models, including Large Language Models (LLMs). However, there is little to no research on the reasons for their poor performance on this task. In this paper, we establish the benchmark results for three popular LLMs: {\tt Gemma2}, {\tt LLaMA3} and {\tt ChatGPT}, showing that their performance on this task is still significantly below that of humans. We also investigate why these models struggle to achieve superior performance. We release our code and introduced datasets at \url{ https://github.com/bodasadallah/decrypting-crosswords}.
\end{abstract}


\section{Introduction}
\label{sec:intro}
\begin{figure}[t!]
\centering
\includegraphics[width=.4\textwidth]{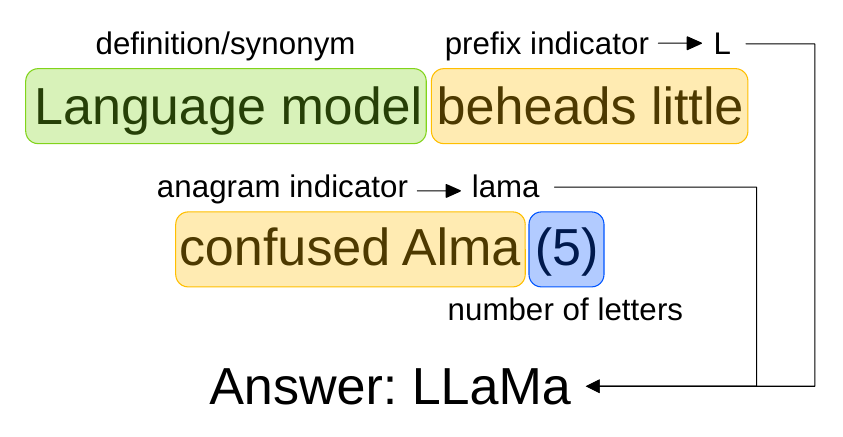}
\caption{\label{fig:example_clue} An example of a cryptic clue: number 5 at the end of the clue denotes the number of characters in the answer and is called {\bf enumeration}. The {\bf definition} part here is \textit{language model}, with the rest being the \textbf{wordplay} part. \textit{Beheads} or similar words point to the first letters of the next word, while \textit{confused} (as well as \textit{mixed up}, etc.) is likely to indicate an anagram. As we should look for a language model's name that starts with the letter \textit{l} plus an anagram of \textit{Alma} and consists of 5 letters, the answer here is \textit{LLaMA}.}
\label{fig:example}
\end{figure}

A cryptic crossword is a type of crossword puzzle known for its enigmatic clues~\citep{crosswords-definition}. Unlike standard crossword puzzles, where clues are straightforward definitions or synonyms of the answers, cryptic crosswords involve wordplay, riddles, and cleverly disguised hints that make solving them more challenging~\citep{collinsbook}. Figure \ref{fig:example} shows an example of such a puzzle.

To solve a cryptic clue, one must not only apply generic rules in the specific context of the clue but also use domain-specific knowledge to produce a reasonable answer. Therefore, tackling cryptic crosswords with modern NLP methods provides a novel and interesting challenge. It has been shown that NLP models' performance is far from that of humans:~\citet{rozner2021decrypting} and ~\citet{efrat2021cryptonite} report an accuracy of 7.3\% and 8.6\% for rule- and transformer-based models. \citet{sadallah2024llmsgoodcrypticcrossword} and~\citet{saha2024languagemodelscrosswordsolvers} show similarly low results for LLMs. In contrast, expert human solvers achieve 99\% accuracy and self-proclaimed amateurs reach 74\%~\citep{friedlander2009expertise, friedlander2020fluid}, however, there are still no official statistics for average human performance.

Typically, a cryptic clue can be divided into two parts: the \textbf{definition} and the \textbf{wordplay} (see Figure \ref{fig:example}). The definition consists of one or more words in the clue that can be used interchangeably with the answer, and it usually appears either at the beginning or at the end of the clue. The wordplay can take many forms: the most popular ones include {\em anagrams}, {\em hidden words}, and {\em double definitions}, among others (see Table \ref{tab:wordplay_types} for popular wordplay types and their examples).

Past approaches to solving cryptic clues range from rule-based models to traditional machine learning models like KNN \citep{rozner2021decrypting} and transformers like T5 \citep{rozner2021decrypting, efrat2021cryptonite}. However, all these models achieve only modest accuracy on the task (see Section \ref{sec:related-work}).
The fact that LLMs can develop emergent capabilities~\cite{wei2022emergent} suggests that they may be able to solve cryptic puzzles if not on a par with human solvers, then at least somewhat successfully, however, our preliminary investigation shows that a zero-shot, naive approach to evaluating LLMs yields very low accuracy. Recently,~\citet{sadallah2024llmsgoodcrypticcrossword} and~\citet{saha2024languagemodelscrosswordsolvers} have evaluated modern LLMs on the task and showed that using certain prompting techniques can help push the limits of LLMs on this task, yet they are still far behind human experts' performance.

In this work, we focus on the interpretability of the LLM performance on the task of cryptic crossword solving and analyze which aspects of the task cause models to struggle most. Therefore, we evaluate LLMs solving one cryptic clue at a time rather than in a grid-eliminating information flow from the rest of the grid, in contrast to previous work like that of~\citet{saha2024languagemodelscrosswordsolvers}. We focus on three main areas of the models' reasoning: (1) we explore whether they can extract the definition part of the clue; (2) we test the models' ability to identify the wordplay type in prompts containing varying amount of information; and (3) we test the models' internal reasoning by asking them to explain how they arrived at the answer.

Our main contributions are as follows: {\bf (1)} We explore the general abilities of LLMs on the challenging task of solving cryptic crosswords using simple prompting strategies, each with a different amount of information embedded into the prompts; {\bf (2)} We investigate models' understanding of the task by addressing three auxiliary tasks; {\bf (3)} To facilitate reproducibility of our results and follow-up experiments, we introduce a small new dataset annotated with wordplay type labels,\footnote{\url{https://huggingface.co/datasets/boda/small_explanatory_dataset}} and release all our data and code.\footnote{\url{https://github.com/bodasadallah/decrypting-crosswords}}

\section{Related Work}

\label{sec:related-work}

Although prior work looked into wordplay \citep{luo2019pungan, he-etal-2019-pun, joker} and traditional crosswords \citep{LITTMAN200223, zugarini2023ratselrevolution}, much less attention has been paid to cryptic crosswords. The early work of \citet{rule_based_solver} achieved $7.3\%$ accuracy on the task with a rule-based solver,\footnote{\url{https://github.com/rdeits/cryptics}} which applied hand-crafted probabilistic context-free grammar to generate all possible syntactic structures for clue words.
Following this, \citet{efrat2021cryptonite} introduced {\tt Cryptonite}, a dataset of 523,114 cryptic clues collected from {\em The Times} and {\em The Telegraph}.  
They fine-tuned a T5~\citep{t5} model, which helped set the benchmark accuracy for Transformer models at $7.6\%$. Similarly, \citet{rozner2021decrypting} introduced a dataset extracted from \textit{The Guardian} and used a curriculum learning approach~\citep{Soviany2021CurriculumLA}, which involved training a model on simpler tasks before progressing to more complex compositional clues. This increased the performance to $21.8\%$.

Recently,~\citet{sadallah2024llmsgoodcrypticcrossword} and ~\citet{saha2024languagemodelscrosswordsolvers} have evaluated multiple LLMs on the task of solving cryptic crossword puzzles using a range of prompting techniques, including zero and few-shot learning. While~\citet{sadallah2024llmsgoodcrypticcrossword} also explicitly fine-tune open-source LLMs for this task,~\citet{saha2024languagemodelscrosswordsolvers} use a combination of chain-of-thought (CoT)~\citep{wei2023chainofthought} and self-consistency (SC)~\citep{wang2023selfconsistency} techniques, and achieve an accuracy score of $20.85\%$ with GPT4-turbo~\citep{openai2024gpt4technicalreport}.
Both conclude that LLMs' performance on this task is still far from that of human experts. However, neither work further analyzes the models' behavior or why they struggle with this task.


\section{Data}
\subsection{{\em The Guardian} dataset}
In our experiments, we primarily use the dataset introduced by \citet{rozner2021decrypting}, which was extracted from \textit{The Guardian}. Most previous models were tested on this dataset, so we have chosen it for comparison purposes as well. In total, the dataset contains 142,380 clues.~\citet{rozner2021decrypting} introduced two different splits for it: {\em naive (random)} and {\em word-initial disjoint}. We evaluate our models on the test subset of 28,476 examples from the {\em naive (random)} split, as it has more diverse examples than the other split.

\subsection{{\em Times for the Times} dataset}
\label{additional-data}
To test models' performance across datasets, we use the data collected by George Ho,\footnote{\url{https://cryptics.georgeho.org/}} where every clue has a marked definition. The original dataset contains around 600k clues from many sources, which would result in extremely expensive experimentation with LLMs. For our experiments, we have sampled 1,000 representative examples\footnote{\url{https://huggingface.co/datasets/boda/times_for_the_times_sampled}} collected from the {\em Times for the Times} blog.\footnote{\url{https://times-xwd-times.livejournal.com/}} We ensure that the distribution of these examples, with respect to the number of words in the definition and their position in the clue, is similar to the full dataset and rely on the available definitions to estimate how well our models understand what the definition is. Additionally, this information helps investigate whether including the definition explicitly aids the models in solving the clues.

\subsection{Small explanatory dataset}
Unfortunately, there is no large-scale dataset that contains information about the wordplay types of the clues. To investigate whether our models can detect wordplay types, we have annotated 200 examples from the additional dataset (see Section \ref{additional-data}), including 40 clues for each major wordplay type ({\em anagram}, {\em assemblage}, {\em container}, {\em hidden word}, and {\em double definition} – see Table \ref{tab:wordplay_types} for examples).

\section{Methodology}

\subsection{Zero-shot setup}
\paragraph{Base prompt}
We begin by defining a simple prompt (see Figure \ref{fig:base_prompt}) that only includes the minimal information required to solve the task. We include the line "you are a cryptic crosswords expert", as it has been shown that this phrase can help the model performance~\citep{xu2023expertprompting}.


\paragraph{All-inclusive prompt}
In this prompt, we combine general information about cryptic crossword solving without adding examples or CoT~\citep{wei2023chainofthought} (see Figure \ref{fig:all_inclusive_prompt}). We include information about clue parts and their meanings. We also add information about the typical position of the definition in the clue. Finally, our preliminary experiments suggest that LLMs often struggle to understand the constraints of the answer length mentioned in the clue, so we explicitly tell the model that the number of letters in the answer is indicated in parentheses at the end of the clue. In addition,  we experiment with solving a cryptic clue using the definition provided.

\subsection{Dividing solution process into sub-tasks}
Next, we investigate why the models struggle to solve the task. To do that, we design experiments to test the models' ability to (1) extract definition word(s) from the clue, (2) detect the wordplay type with varying levels of information, and (3) explain the solution process given the clue and the answer.

\section{Experiments and Discussion}
We choose two of the most recent and popular open-source LLMs, {\tt Gemma2}~\citep{gemmateam2024gemmaopenmodelsbased} and {\tt LLaMA3}~\citep{llama3modelcard}, and one closed-source model, {\tt ChatGPT}~\citep{chatgpt}. The details are provided in Appendix \ref{hyperparameters_appendix}, and the results in Table \ref{experiments_results}.

\subsection{Cryptic clue solving}
The first four rows of Table \ref{experiments_results} show the models' accuracy in solving cryptic clues on two different datasets for two different prompts. We can see that {\tt ChatGPT} outperforms the open-source models. Also, we can conclude that providing the models with the definition improves their performance. To put these results into perspective, in Table~\ref{final_comp}, we compare our results with those obtained by~\citet{rozner2021decrypting}. We do not compare to the results from ~\citet{saha2024languagemodelscrosswordsolvers} because they are reported on a different subset of the dataset from ~\citet{rozner2021decrypting}. We observe that using {\tt ChatGPT} in a zero-shot setting achieves results comparable to (but still lower than) those of T5 fine-tuning. One important thing to note is that~\citet{rozner2021decrypting} explicitly fine-tuned models on the task, while the models we used are general LLMs that were pre-trained on the generic language modeling task.

\subsection{Understanding various aspects of the task}
\subsubsection{Definition extraction}
We ask the models to extract the definition part of the clue with the prompt illustrated in Figure \ref{fig:definition_prompt}. We specify that the definition should be a synonym for the answer but do not indicate that the definition usually appears at the beginning or end of the clue.
All models show higher results in the definition extraction task. One reason for this could be that the definition is explicitly included in the clue, making the task a matter of repeating part of the clue, which is generally easier than generating new words as an answer.

\begin{table*}
\centering
\begin{tabular}{p{3.84cm}p{1.85cm}p{3.8cm}ccc}
\toprule
& Number & & \multicolumn{3}{c}{Accuracy} \\
Task & of examples & Info / Prompt & {\tt LLaMA3} & {\tt ChatGPT}  & {\tt Gemma2}\\
\cmidrule(rl){1-6}
Cryptic Clue Solution & 28476 & base prompt & 2.2 & 10.9 & 4.8\\
Cryptic Clue Solution & 28476 & all inclusive prompt & 2.1 & \textbf{11.4} & 2.4 \\
Cryptic Clue Solution & 1000 & all inclusive prompt & 3.3 & 13.4  & 5.3\\
Cryptic Clue Solution & 1000 & $ \sim$ + definition & 3.8 & {\bf 16.2}  & 7.0\\
\cmidrule(rl){1-6}
Definition Extraction & 1000 & definition extraction& 19.3 & {\bf 41.2} & 21.8\\
\cmidrule(rl){1-6}
Wordplay Type Detection & 200 & wordplay types &  20.0 & 42.5   & 33.5\\
Wordplay Type Detection & 200 & $\sim$ + explanation + ex. & 23.0 & 43.5 & 39.0 \\
Wordplay Type Detection & 200 & $\sim$ + clue answer & 23.5 &  {\bf 44.5} & 43.5\\
\bottomrule
\end{tabular}
\caption{\label{experiments_results}The summary of the results obtained in our experiments on the {\tt naive (random)} (first two rows), {\tt Times for the Times} (rows 3-5), and {\tt small explanatory} (last three rows) datasets. Best results are highlighted in bold.}
\end{table*}


\subsubsection{Wordplay detection}
\paragraph{Determining the wordplay type}
We identify five major types of wordplay listed in Table \ref{tab:wordplay_types}. Then we investigate if our models could identify the wordplay type from the clues. Usually, professional solvers note indicator words that relate the clue to one type or another: for example, \textit{confused, mixed up}, and {\em mad} usually indicate anagrams.
To test the models' ability to identify the wordplay type, we design three experiments that gradually add information to the prompt. The specific design of the experiments is described in the Appendix \ref{wordplay_types_experiments_apendix}.

The results show that adding the definition for the wordplay and providing a model with the answer do not significantly improve the model's ability to extract the wordplay type except for {\tt Gemma}, which has a performance increase of 10\%. {\tt LLaMA3} only predicted one wordplay type ({\em hidden word}) using the `wordplay types' prompt (see Figure~\ref{fig:wordplay_types_simplest}), but providing more information in the other prompts helped the model predict other types. We hypothesize that a potential reason for {\tt LLaMA3}'s behavior is that the model seems to attend more to the task prompt than the clue itself.

We acknowledge that the small dataset size might constrain our ability to draw definitive conclusions. However, an important observation is that all 3 models over-predict some types ({\em anagram} and {\em hidden word}) while under-predicting others ({\em assemblage}). We include the full analysis with the models' confusion matrices on the most informative prompt shown in Figure~\ref{fig:wordplay_types_examples} in Appendix~\ref{wordplay_types_experiments_apendix}.

\subsubsection{Explanation extraction}

Finally, we ask the models to explain the solution, given the clue and the answer. Our analysis of the models' answers shows that: (1) All the models follow some kind of structure in their explanations, breaking the clue into parts of one to three words; however, this separation often does not seem to make sense, as it may combine both definition and wordplay parts together or use words that do not interact with each other. (2) {\tt LLaMA3} does not mention any wordplay operations and only works at a synonym level, which is insufficient for solving the clues. (3) {\tt Gemma} shows the knowledge of some operation types (such as anagram and even homophones-related operations) but applies it incorrectly. (4) {\tt ChatGPT} recognizes that something should be done with the characters and words in the clue and sometimes even gets it right, for example, suggesting taking an anagram of a given word or putting together words in an assemblage clue; however, it does not properly "understand" the procedure. For instance, one of the {\tt ChatGPT}'s outputs is: \textit{rearranging the letters of "pan" and adding "to cook cheese" results in "parmesan"}. This statement is incorrect, as one cannot get "parmesan" from the letters in "pan" and "to cook cheese." (5) The easiest type to generate sensible explanations for are clues for the {\em double definition} type, where both parts of the clue are synonymous with the answer – this aligns with how base LLMs were trained.

\begin{table}
\centering
\begin{tabular}{cc}
\toprule
Model & Accuracy  \\
\cmidrule(rl){1-2}
{\tt LLaMA3} (best) & 2.2 \\
{\tt Gemma2} (best) & 4.8\\
{\tt ChatGPT} (best) & 11.4 \\
\cmidrule(rl){1-2}
Rule-based & 7.3 \\

T5 fine-tuned&  16.3 \\
T5 fine-tuned + curriculum &  21.8 \\
\bottomrule
\end{tabular}
\caption{\label{final_comp}Comparison with previous results: a rule-based method of~\citet{rule_based_solver} and the T5-based approach of~\citet{rozner2021decrypting}.}
\end{table}

\section{Conclusions and Future Work}



In this work, we have focused on studying the inner workings of LLMs while solving cryptic crosswords rather than trying to improve their performance on this task. We began by evaluating the models under a zero-shot setting and then tried to gain insights into their understanding of cryptic clues through auxiliary tasks. The results suggest that although {\tt ChatGPT} model overall outperforms open-source LLMs, solving cryptic crosswords remains a very challenging task for all tested LLMs, with a significant room for improvement. In addition, we conclude that splitting the task into subtasks helps the models to some extent, which indicates that models cannot break down the composite task by themselves. The performance of the models on the chosen subtasks still remains unsatisfactory: the models struggle to identify the definition and the wordplay type.

We believe the performance can be improved in future work using several possible research directions. Firstly, promising avenues for research in this area are chain-of-thought~\cite{wei2023chainofthought} and tree-of-thought~\cite{yao2023tree} techniques. This is motivated by our current results that suggest that splitting the task into simpler subtasks helps improve the model performance: specifically, CoT-based methods can teach models how to arrive at the solution step-by-step by splitting the original complex task into such multiple simpler subtasks. 
Secondly, given the considerable performance increase achieved using curriculum learning with T5~\cite{rozner2021decrypting}, we consider this direction worth exploring with LLMs as well. Finally, approaches such as a mixture of experts~\citep{moe_1991,moe_megablock} used to train open-source models like Mixtral~\citep{jiang2024mixtral} can be applied to the task, as models may develop expert layers specializing in separate wordplay types.

\section*{Limitations}

\paragraph{Limited set of LLMs experimented with} Experiments with an extensive set of state-of-the-art LLMs can get quite expensive. Due to budget limitations, we have been selective in terms of the LLMs that we use in this study. Specifically, we chose only a few of the most popular open-source and closed-source LLMs. We believe that the obtained results shed light on the current LLMs' capabilities on this task. However, we acknowledge that the set of LLMs we tested here is limited, and our results cannot be extrapolated to other LLMs. In addition, in many experiments, we have observed that minor changes in settings do not bring substantial improvement to the results. This motivated us to perform only a limited set of experiments with the chosen models, as elaborated in the paper.

\paragraph{Limitations of the datasets size} Some datasets we used are not large in terms of the number of examples. The main reason for this is the lack of existing datasets with rich annotation, so we had to create one such dataset ourselves. We acknowledge that the results obtained on a larger dataset may be more reliable; however, we believe that the results reported here already provide us with useful insights.


\paragraph{Closeness to the real-world scenario}
In this work, we have focused on solving one clue at a time. In the real-world scenario, human solvers encounter twenty to thirty clues in one grid. Solving one clue usually reveals letters of the other answers, which can be quite helpful in the solution process. In contrast, our goal is to investigate LLMs' abilities in cryptic crossword clue interpretation, and we do not try to solve the whole grid.

\paragraph{Dangers of data contamination} Finally, we observe in our experiments that {\tt ChatGPT} outperforms the open-source models. We acknowledge that we lack information about its training setup, as {\tt ChatGPT} is a proprietary model, and therefore, we cannot guarantee that this model's training data is uncontaminated; in other words, it is not entirely clear whether the model could have been exposed to any of the crossword clues during its training. 
However, we note that all LLMs still struggle to solve cryptic clues, showing that even if some contamination took place, the models do not seem to be able to memorize and simply reproduce the answers from previously seen clues. As a side note, human experts also get exposed to a lot of clues in their practice, and their performance on the task is still much higher than that of LLMs.


\section*{Ethics Statement}
We foresee no serious ethical implications from this study.

\section*{Acknowledgments}
\label{sec:ack}
We are grateful to the campus supercomputing center at MBZUAI for providing resources for this research.

\bibliography{references}

\appendix

\section{Implementation details}\label{hyperparameters_appendix}
For the open-source models, we chose to use the instruct-tuned versions ({\tt gemma-2-9b-it} and {\tt Meta-Llama-3-8B-Instruct}) as they give better results and are more suitable for the task. For all three models, we generated the outputs using zero temperature (greedy sampling) as it showed slightly better results in our preliminary experiments. To get the highest possible performance from the open-source models, we used the models in full precision, without doing any quantization. Finally, we used FlashAttention-2~\citep{dao2023flashattention2fasterattentionbetter} for faster inference.

We used {\tt gpt3.5-turbo} with near-zero temperature and $top\_p$ of 1e-9. However, as it gave slightly different results across different runs, we ran it 3 times and reported the average of the results from these runs. 

We do not use any post-processing of the models' answers.

\section{Wordplay types}\label{wordplay_types_appendix}
\setcounter{table}{0}
\renewcommand\thetable{B\arabic{table}}

Common wordplay types are listed in Table \ref{tab:wordplay_types} with examples\footnote{Examples are taken from \url{https://crypticshewrote.wordpress.com/explanations/}} and explanations. We identify 5 main types: anagram, assemblage, container, hidden word, and double definition.
\begin{table*}
\centering
\begin{tabular}{p{7cm}p{6cm}c} 
\toprule
Type        & Example Clue & Answer   \\
\cmidrule(r){1-3}
\textbf{Anagram:} certain words or letters must be jumbled to form an entirely new term. & \underline{Never} upset \textbf{a Sci Fi writer} (5)  &  Verne \\
\textbf{Assemblage:} the answer is broken into its component parts and the hint makes references to these in a sequence. & \textbf{Bitter} initially, \underline{b}ut \underline{e}xtremely \underline{e}njoyable \underline{r}efreshment  (4)  & Beer \\
\textbf{Container:} the answer is broken down into different parts, with one part embedded within another. & \textbf{The family member} put \underline{us} in the \underline{money} (6)  & Cousin \\
\textbf{Hidden word:} the answer will be hidden within one or multiple words within the provided phrase. & \textbf{Confront them} in the tob\underline{acco st}ore (6) & Accost \\
\textbf{Double definition:} contains two meanings of the same word. & \textbf{In which you’d place the photo} of the \textbf{NZ author} (5)  & Frame \\
\bottomrule
\end{tabular}
\caption{Examples of common wordplay types. The definition part is bolded.}
\label{tab:wordplay_types}
\end{table*}

\section{Worplay type detection experiments}\label{wordplay_types_experiments_apendix}
\setcounter{figure}{0}
\renewcommand\thefigure{C\arabic{figure}}
In the first experiment, we give the models the names of the five different wordplay types and ask them to predict which wordplay type the given clue belongs to (see Figure \ref{fig:wordplay_types_simplest}). We notice that {\tt LLaMA3} fails to understand the task and produces only one type for all examples, which suggests that the model does not analyze the given clues thoroughly. Next, we experiment with providing the models with the explanations and one example for each wordplay type (Figure \ref{fig:wordplay_types_examples}). Finally, we add the answer for each clue to test whether the models can infer information about the wordplay types from the answer (Figure \ref{fig:wordplay_types_examples_answer}).

\begin{figure}
    \centering
    \includegraphics[width=0.48\textwidth]{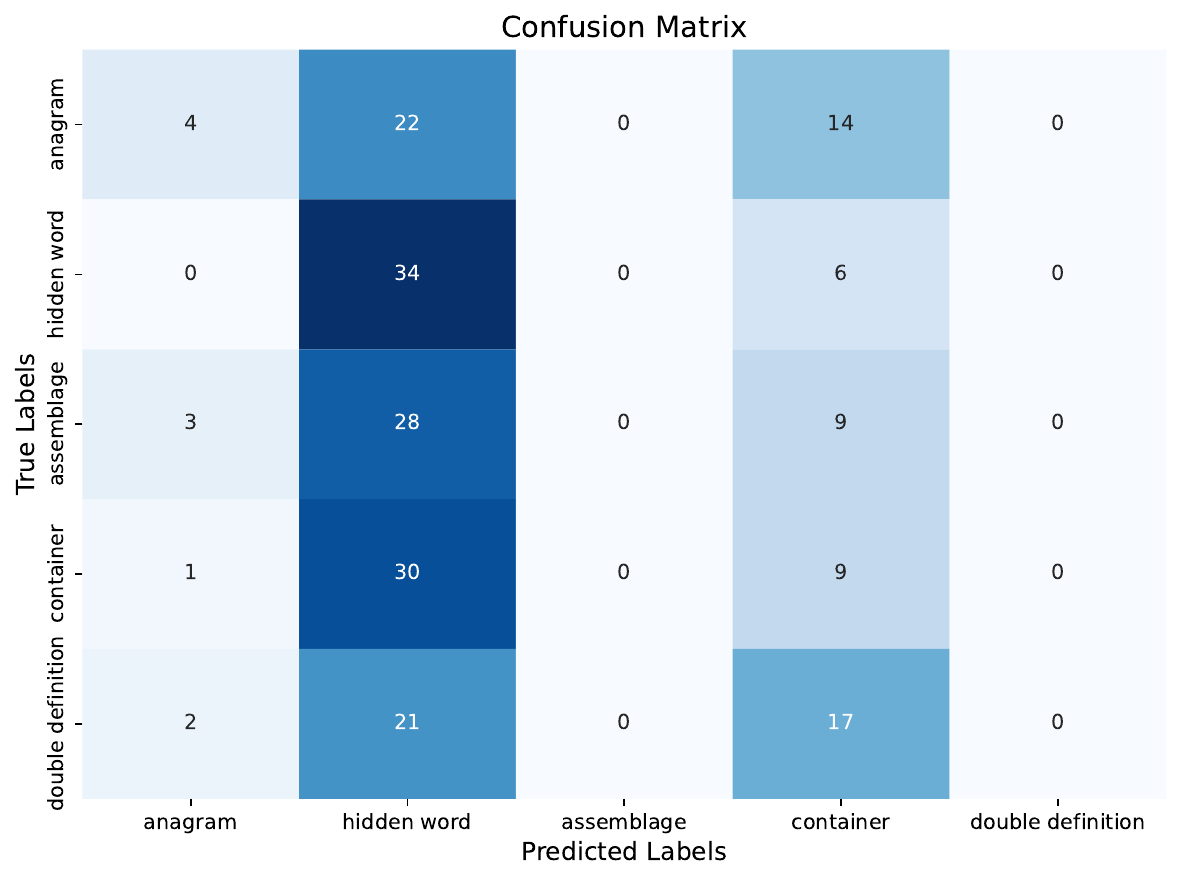}
    \caption{Confusion matrix for {\tt LLaMA3} on wordplay type prediction using the most informative prompt~\ref{fig:wordplay_types_examples_answer}.}
    \label{fig:confusion_llama}
\end{figure}

Next, we analyze the models' predictions using the most informative prompt (Figure~\ref{fig:wordplay_types_examples_answer}).
For {\tt LLaMA3}, the most frequently predicted  wordplay type is "hidden word" (100+ samples) and "container" (55 samples), and never predicted "double definition" or "assemblage." The confusion matrix is shown in Figure~\ref{fig:confusion_llama}.

\begin{figure}
    \centering
    \includegraphics[width=0.48\textwidth]{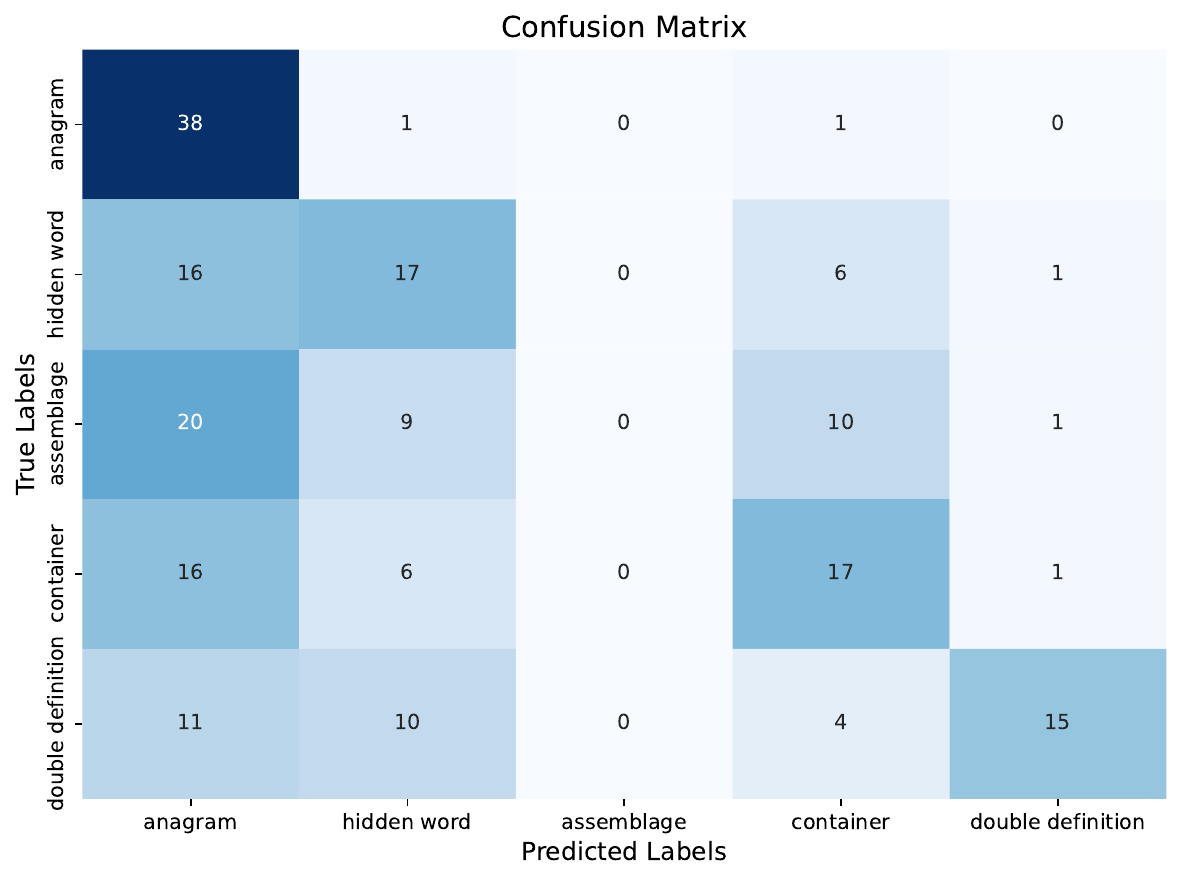}
    \caption{Confusion matrix for {\tt Gemma} on wordplay type prediction using the most informative prompt~\ref{fig:wordplay_types_examples_answer}.}
    \label{fig:confusion_gemma}
\end{figure}

{\tt Gemma} most frequently predicted "anagram" (101 samples) and "hidden word" (43 times) and never predicted "assemblage." Its confusion matrix is shown in Figure~\ref{fig:confusion_gemma}.

\begin{figure}
    \centering
    \includegraphics[width=0.48\textwidth]{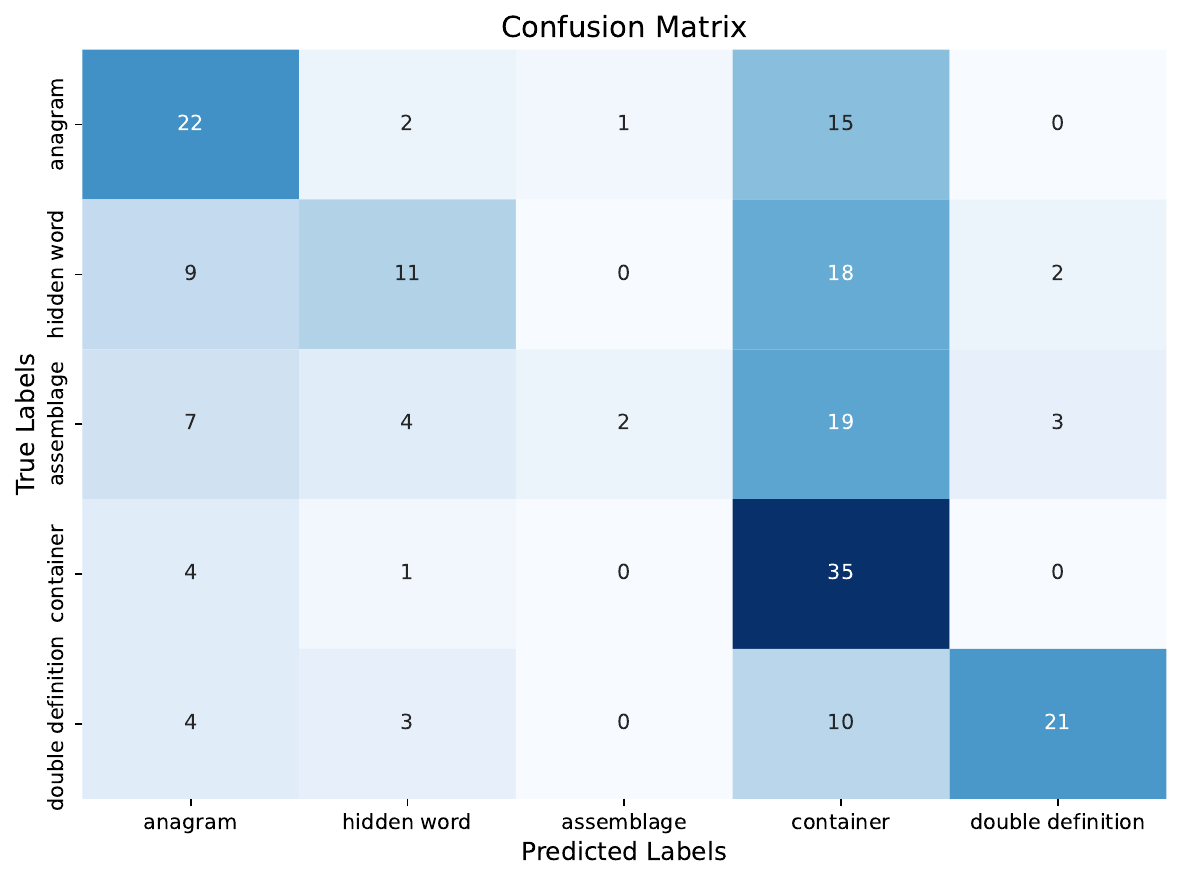}
    \caption{Confusion matrix for {\tt ChatGPT} on wordplay type prediction using the most informative prompt~\ref{fig:wordplay_types_examples_answer}.}
    \label{fig:confusion_chatgpt}
\end{figure}

{\tt ChatGPT} most frequently predicted "container" (97 times) and "anagram" (46 times) and predicted "assemblage" only 3 times. Its confusion matrix is shown in Figure~\ref{fig:confusion_chatgpt}. What is interesting here is that the model sometimes predicted types different from the specified ones.


\section{Data sources}
In the text of the paper, we mention several sources of cryptic crosswords: 
\begin{enumerate} \vspace{-0.5em}
    \item \textit{The Times}\footnote{\url{https://www.thetimes.co.uk/puzzleclub/crosswordclub/home/crossword-cryptic}} \vspace{-0.5em}
    \item \textit{Telegraph}\footnote{\url{https://puzzles.telegraph.co.uk/crossword-puzzles/cryptic-crossword}} \vspace{-0.5em}
    \item \textit{The Guardian}\footnote{\url{https://www.theguardian.com/crosswords/series/cryptic}} \vspace{-0.5em}
    \item \textit{Times for the Times} blog\footnote{\url{https://times-xwd-times.livejournal.com/}}\vspace{-0.5em} 
\end{enumerate}
We do not parse their data specifically but rather use already prepared datasets or samples from them.

\section{Prompts}
\setcounter{figure}{0}
\renewcommand\thefigure{E\arabic{figure}}
\label{prompts_appendix}
We present all the prompts we used in this section: see Figures \ref{fig:base_prompt} to \ref{fig:wordplay_types_examples_answer}.
\begin{figure}[!h]
    \centering
    \includegraphics[width=0.42\textwidth]{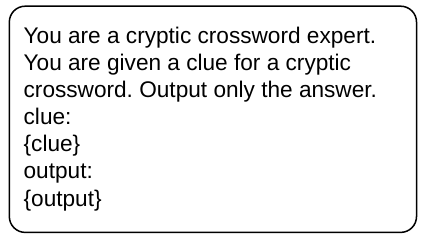}
    \caption{Base prompt.}
    \label{fig:base_prompt}
\end{figure}

\begin{figure}[!h]
    \centering
    \includegraphics[width=0.42\textwidth]{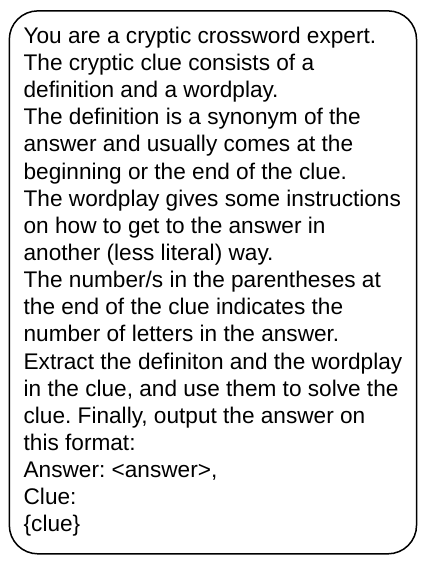}
    \caption{All inclusive prompt.}
    \label{fig:all_inclusive_prompt}
\end{figure}

\begin{figure}[!h]
    \centering
    \includegraphics[width=0.42\textwidth]{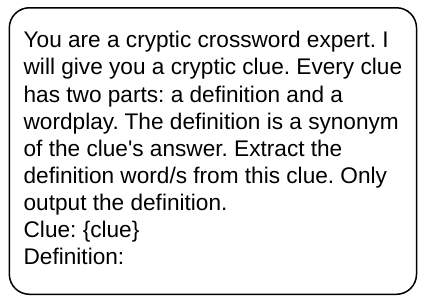}
    \caption{Prompt for the definition extraction.}
    \label{fig:definition_prompt}
\end{figure}

\begin{figure}[!h]
    \centering
    \includegraphics[width=0.42\textwidth]{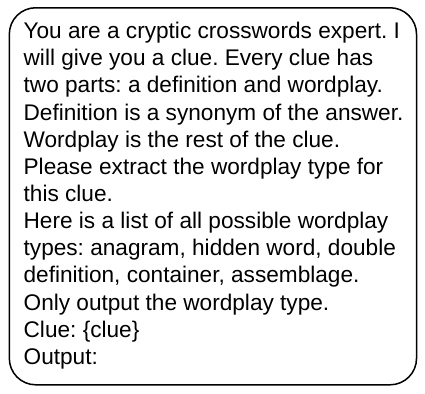}
    \caption{Prompt for the wordplay type classification.}
    \label{fig:wordplay_types_simplest}
\end{figure}

\begin{figure}
    \centering
    \includegraphics[width=0.42\textwidth]{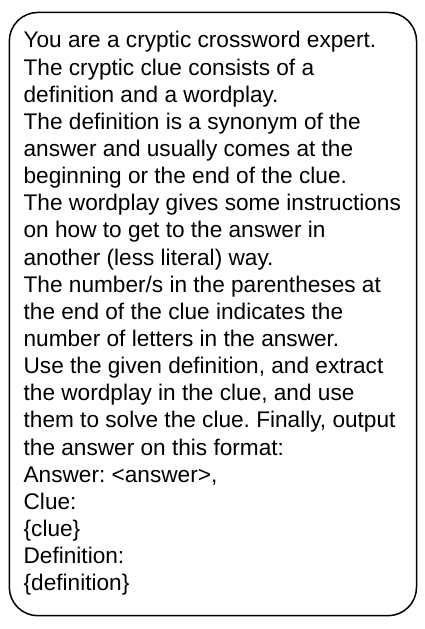}
    \caption{All inclusive prompt with included definition.}
    \label{fig:solution_with_definition}
\end{figure}

\begin{figure*}
    \centering
    \includegraphics[width=0.98\textwidth]{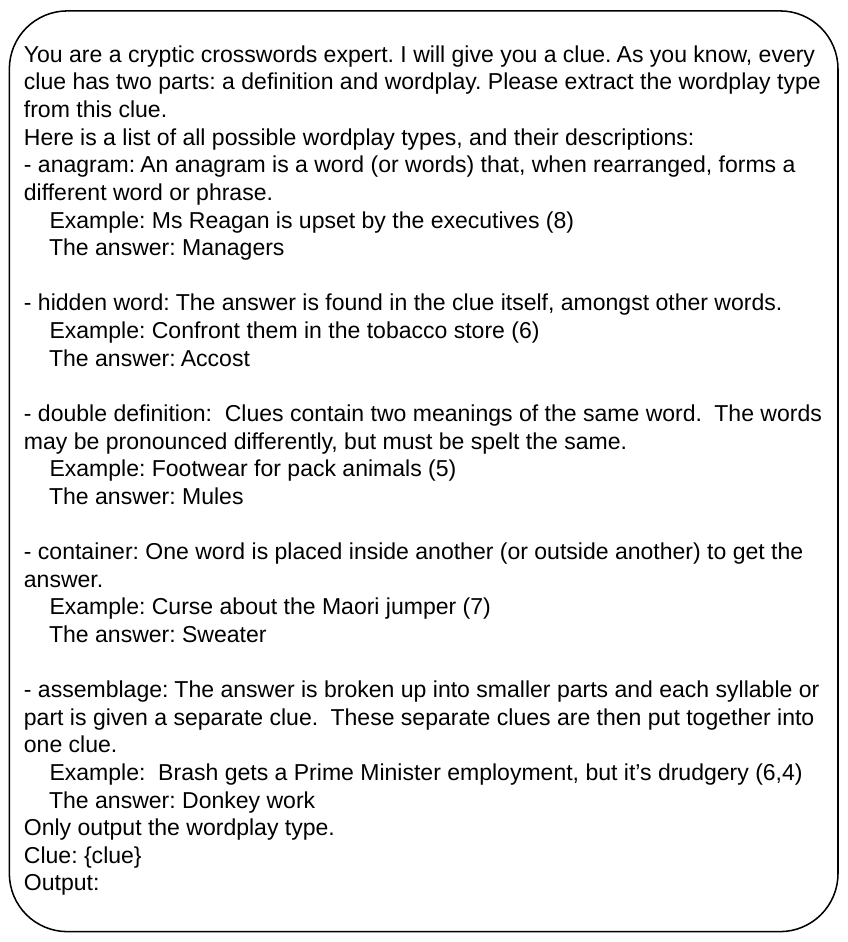}
    \caption{Prompt for the wordplay type classification with examples for each wordplay type.}
    \label{fig:wordplay_types_examples}
\end{figure*}

\begin{figure*}
    \centering
    \includegraphics[width=0.98\textwidth]{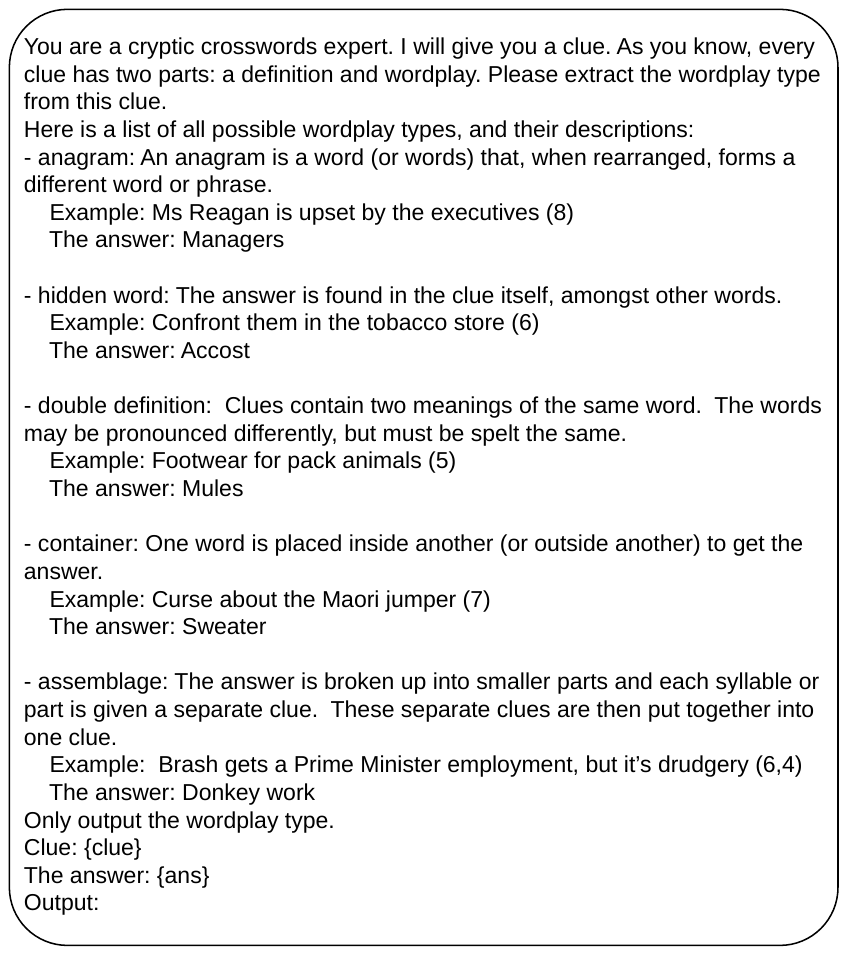}
    \caption{Prompt for the wordplay type classification with examples for each wordplay type. Here we also add the answer for the clue.}
    \label{fig:wordplay_types_examples_answer}
\end{figure*}

\end{document}